\documentclass[conference]{IEEEtran}
\IEEEoverridecommandlockouts
\usepackage{cite}
\usepackage{amsmath,amssymb,amsfonts}
\usepackage{graphicx}
\usepackage{textcomp}
\usepackage{xcolor}
\usepackage{hyperref}
\usepackage[numbers]{natbib}

\usepackage{amsmath}
\usepackage{wrapfig}
\usepackage{algorithm}
\usepackage[noend]{algorithmic}
\usepackage{subcaption}
\usepackage{booktabs}
\usepackage{soul}
\usepackage{multirow}
\usepackage{amsthm}
\theoremstyle{definition}
\newtheorem{definition}{Definition}[section]
\usepackage{bm}

\def\BibTeX{{\rm B\kern-.05em{\sc i\kern-.025em b}\kern-.08em
    T\kern-.1667em\lower.7ex\hbox{E}\kern-.125emX}}
\begin{document}

\title{Higher Replay Ratio Empowers Sample-Efficient Multi-Agent Reinforcement Learning}

\author{
  Linjie Xu\textsuperscript{1}, Zichuan Liu\textsuperscript{2}, Alexander Dockhorn\textsuperscript{3}, Diego Perez-Liebana\textsuperscript{1},\\
  Jinyu Wang\textsuperscript{4},  Lei Song\textsuperscript{4}, Jiang Bian\textsuperscript{4}\\
  \textsuperscript{1}Queen Mary University of London, \textsuperscript{2}Nanjing University,\\
  \textsuperscript{3}Leibniz University Hannover,  \textsuperscript{4}Microsoft Research Asia \\
  \texttt{\{linjie.xu, diego.perez\}@qmul.ac.uk}, \\
  \texttt{zichuanliu@smail.nju.edu.cn},
  \texttt{dockhorn@tnt.uni-hannover.de}, \\
 \texttt{\{wang.jinyu, lei.song, jiang.bian\}@microsoft.com}
}

\maketitle

\begin{abstract}
One of the notorious issues for Reinforcement Learning (RL) is poor sample efficiency. Compared to single agent RL, the sample efficiency for Multi-Agent Reinforcement Learning (MARL) is more challenging because of its inherent partial observability, non-stationary training, and enormous strategy space. Although much effort has been devoted to developing new methods and enhancing sample efficiency, we look at the widely used episodic training mechanism. In each training step, tens of frames are collected, but only one gradient step is made. We argue that this episodic training could be a source of poor sample efficiency. To better exploit the data already collected, we propose to increase the frequency of the gradient updates per environment interaction (a.k.a. Replay Ratio or Update-To-Data ratio). To show its generality, we evaluate $3$ MARL methods on $6$ SMAC tasks. The empirical results validate that a higher replay ratio significantly improves the sample efficiency for MARL algorithms. The codes to reimplement the results presented in this paper are open-sourced at \url{https://anonymous.4open.science/r/rr_for_MARL-0D83/}.
\end{abstract}


\section{Introduction}
With the ongoing development of simulation~\citep{macklin2019non, lesser2022loki}, rendering~\citep{dupuy2020concurrent, qiao2023dynamic}, and content generation~\citep{kumaran2023scenecraft, merino2023five, sudhakaran2024mariogpt} techniques, video games have reached a new level of complexity, represented by concurrent events, display resolution, and content diversity. Higher complexity empowers more realistic video games that provide players with a better immersive experience. Meanwhile, it costs significant computing power to run these games. However, when developing agents for these games, these computation becomes an indispensable concern because agent training depends on interacting with the game. Worsely, modern artificial intelligence methods such as Reinforcement Learning (RL), are notorious for their poor sample efficiency and may require hundreds of millions of game interactions~\citep{hessel2018rainbow} for training. In conclusion, the combination of game complexity and poor sample efficiency makes it even more challenging to develop autonomous agents for game playing.

In this work, we focus on the cooperative multi-agent setting, similar to a wide range of games (RTS~\citep{samvelyan2019starcraft}, MOBA~\citep{berner2019dota}, etc). Multi-agent reinforcement Learning (MARL)~\citep{SunehagVDN, rashid2020monotonic, wang2021qplex} is a powerful technique for training agents to play cooperative games. Compared to single agent RL, the sample efficiency for MARL is more challenging because of its inherent partial observability, non-stationary training, and enormous strategy space. Recently, much effort~\citep{SunehagVDN, rashid2020monotonic, wang2021qplex, yu2022surprising, liu2023na2q}, has been devoted to developing new methods and enhancing sample efficiency. However, the training framework remains mostly unchanged. At each training step of MARL, a complete trajectory is collected by taking actions from the current policy. After that, one gradient update step is made to optimize the policy. To converge, current methods usually need to collect millions of trajectories (approximately hundreds of millions of frames), making sample efficiency a notorious issue.

In single agent RL, a line of work~\citep{nikishin2022primacy, d'oro2023sampleefficient, schwarzer2023bigger} shows that increasing the Replay Ratio (RR, a.k.a. Update-To-Data ratio~\citep{chen2021randomized}) significantly accelerates convergence and improves the final performance. This observation provides insight that the collected data can be better exploited by performing more parameter-update steps. Compared to single agent RL, MARL has less frequent parameter updates (per trajectory VS per frame in single agent RL). Therefore, MARL has a higher risk of not exploiting the collected data well. Without learning well about the current data, an agent fails to collect new data efficiently and thus further harms the sample efficiency.

Although the use of higher RRs is studied in single agent RL, to the best of our knowledge, the RR is not yet investigated in MARL. To address the data exploitation issue mentioned above, we propose to increase the RR for MARL training. It is worth noting that there are other approaches to accelerate the parameter updates, such as using higher learning rates or larger batch sizes. Surprisingly, these approaches are much less effective compared to increasing RR (please refer to Section~\ref{sec:lr_bs} for details).

Increasing the RR causes a linear growth in the number of parameter updates. This might raise concerns about losing the neural network plasticity that causes issues such as primacy bias~\citep{nikishin2022primacy}. In this work, we also investigate the plasticity of MARL agents by visualizing the dormant ratio of the neural network during training. Surprisingly, our empirical analysis shows that the RNN (commonly used in MARL) helps maintain plasticity. Therefore, using higher RRs performs well without the need for additional techniques (e.g. reset~\citep{nikishin2022primacy}) to maintain the network plasticity.
\begin{figure*}[h]
    \centering
    \includegraphics[width=0.85\textwidth]{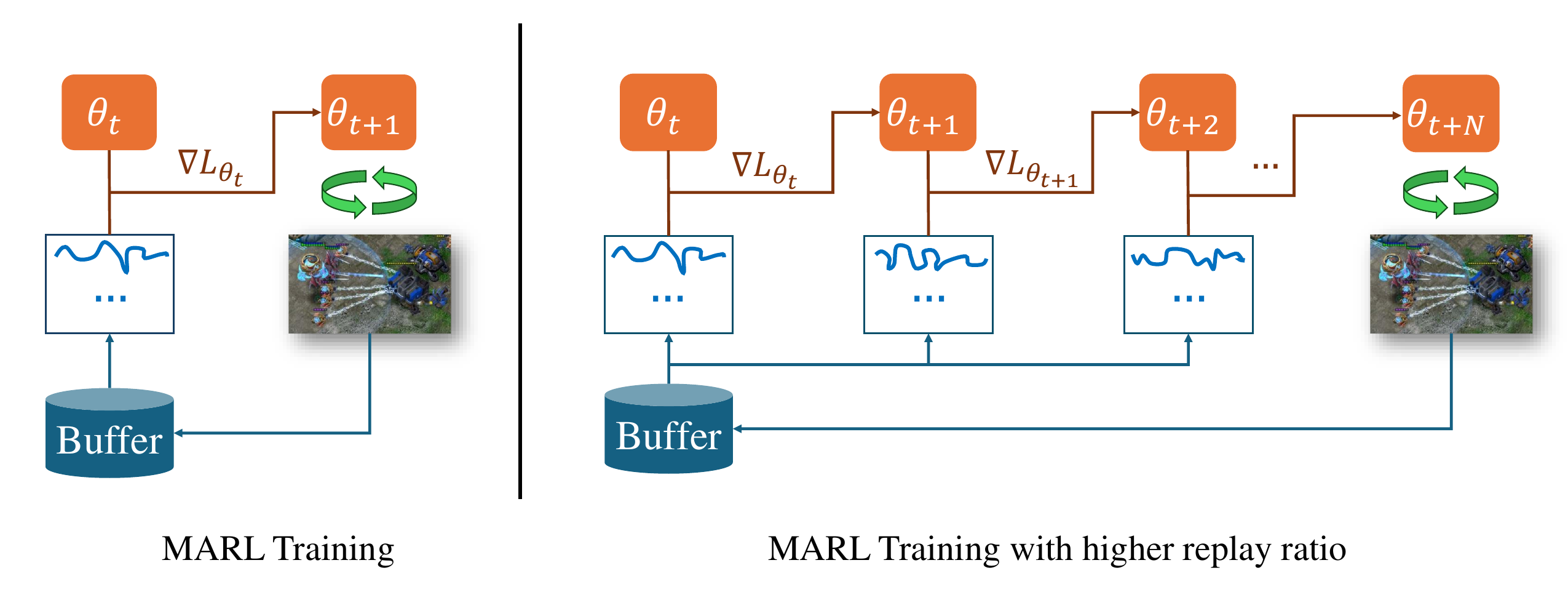}
        \vspace{-2mm}
    \caption{Training pipeline for MARL. $
    \theta_t$ represents the agent parameter on the $t$-th update. For each update, a batch of trajectories is sampled from the replay buffer for the gradient calculation. \textbf{Left:} Conventional MARL training that takes one back-propagation for each environmental interaction (one episode). \textbf{Right:} MARL training with $RR=N$, where multiple backpropagations are applied for each interaction to better exploit the collected data.}
    \label{fig:demo}
\end{figure*}

The contributions of this work are concluded as follows:
\begin{enumerate}
    \item We look into the training mechanism of MARL and propose to increase the RR to enhance sample efficiency.
    \item Our experiments on $3$ widely investigated baselines verify the significance of the proposed method among different StarCraft II tasks.
    \item Although higher RR in single agent RL usually exacerbates the loss of network plasticity, our empirical analysis of dormant neurons reveals a low risk of losing network plasticity for RNN agents used for MARL.
\end{enumerate}

\section{Preliminary}
The cooperative multi-agent setting studied in this work is modeled as a Decentralized Partially Observable Markov Decision Process (Dec-POMDP)~\citep{oliehoek2016concise}. 
A Dec-POMDP is defined as $\mathcal{M} = \langle \mathcal{S}, \mathcal{A}, R, \mathcal{P}, \mathcal{Z}, \mathcal{O}, n, \gamma \rangle$, where $s \in \mathcal{S}$ is the global state. 
Each agent $i \in \{1, ..., n\}$ only receives a partial observation $o_i\in \mathcal{Z}$ from the observation function $\mathcal{O}(s,i)$ to formulate a joint observation  $\mathbf{o}:= [o_i]_{i=1}^n$ at each time step.
These agents choose and execute their own actions $\{a_i \in \mathcal{A} \}_{i=1}^n $ combined as the joint action $\mathbf{a} =[a_i]_{i=1}^n$, and then receive a team reward $r^{tot}$ from the global reward function $R: \mathcal{S} \times \mathcal{A} \mapsto \mathbb{R}$.
This results in a transition to next state $s^{\prime} \sim P\left(s^{\prime} \mid s, \mathbf{a}\right)$, where the transition probability $P\in \mathcal{P}$.
At each training step, an episode of gameplay (a completed trajectory) $\tau$ is collected by interacting with the environment. A trajectory $\tau$ includes a sequence of transitions $T=\{s, \mathbf{o}, \mathbf{a},  r^{tot}, s', \mathbf{o'} \}$, consisting of a global state $s$, a global reward $r^{tot}$, agent observations $\mathbf{o}$, agent actions $\mathbf{a}$, next global state $s'$, and the next agent observations $\mathbf{o'}$.
The goal is to learn a joint policy $\bm{\pi} = [\pi_i]_{i=0}^n$, where $\pi_i: \mathcal{O}_i \mapsto \mathcal{A}_i $, to maximize the expectation of discounted return 
\begin{align}
    G = \mathbb{E}_{s_0 \in \mathcal{S}_0, \mathbf{a_t} \sim \bm{\pi}, s_{t+1} \sim P(s_{t+1}| s_{t}, \mathbf{a}_{t})} \sum_{t=0}^{t<H} \gamma^{t} R(s_t, \mathbf{a}_t),
\end{align}
where  $\gamma \in [0, 1)$ is a discount factor.

\subsection{Centralized-Training Decentralized-Execution}
In the execution of a Dec-POMDP, each agent receives only a local observation. However, it is difficult to make decisions depending on only the local observation that does not reveal complete game information. Fortunately, during training, it is convenient to obtain the global state from the environment. The Centralized-Training Decentralized-Execution (CTDE) is an effective and widely used mechanism that utilizes the global state in agent training. Next, we introduce how CTDE works for two types of RL methods.

In value-based RL methods~\citep{rashid2020monotonic, wang2021qplex, liu2022mixrts}, a mixer function that utilizes the global state is learned with the local agent utility function. The mixer function is used to predict the global value function and is updated by Temporal Difference (TD) learning~\citep{mnih2015human}. In actor-critic methods~\citep{lowe2017multi}, the critic takes the global state as part of the input to better approximate the global value function. The actors, whose input is the local observation, are responsible for the execution.

\subsection{Agent Parameterization and Parameter Update}
If each agent is parameterized independently, the total parameters increase linearly with the number of agents. To obtain a scalable parameterization, a common approach to parameterize agents $\bm{\pi}$ is to use a shared~\citep{SunehagVDN, rashid2020monotonic, li2021celebrating} neural network, which has been shown to outperform independent agents~\citep{PapoudakisC0A21}.

As an effective technique for addressing partial observability, RNN is usually used for MARL agents~\citep{ni22a}. An RNN agent is updated by the backpropagation through a time algorithm mechanism, where the forward passing starts from the beginning of the trajectory and iterates through the whole trajectory. The backpropagation, in reverse, starts from the last time step and ends at the first time step.

\subsection{Data Reuse}
The MARL training iteratively collects data from the environment and updates the agent parameter. The on-policy methods and off-policy methods have different approaches to reuse the data collected at each training step.

The on-policy methods~\citep{yu2022surprising} firstly collect a small-size dataset (e.g. 3,200 frames in Yu et al.~\citep{yu2022surprising}) and then trains on these data for several epochs. After every training step, the dataset is thrown away.
The data reuse for off-policy methods differs in that a replay buffer is maintained. At each training step, an episode is collected and put in a replay buffer. After that, a batch of trajectories $\mathcal{B} = \{\tau_0, \tau_1,...\tau_N \}$ of batch size $N$ are sampled from the replay buffer $\mathcal{D}$.

\section{Methods}
The RR measures the number of update steps per collected data (per frame in single agent RL). In MARL, the update interval is an episode instead of a frame, hence we define the RR for MARL as \textit{the number of gradient steps per episode}.
\begin{figure}
    \centering
    \includegraphics[width=0.42\textwidth]{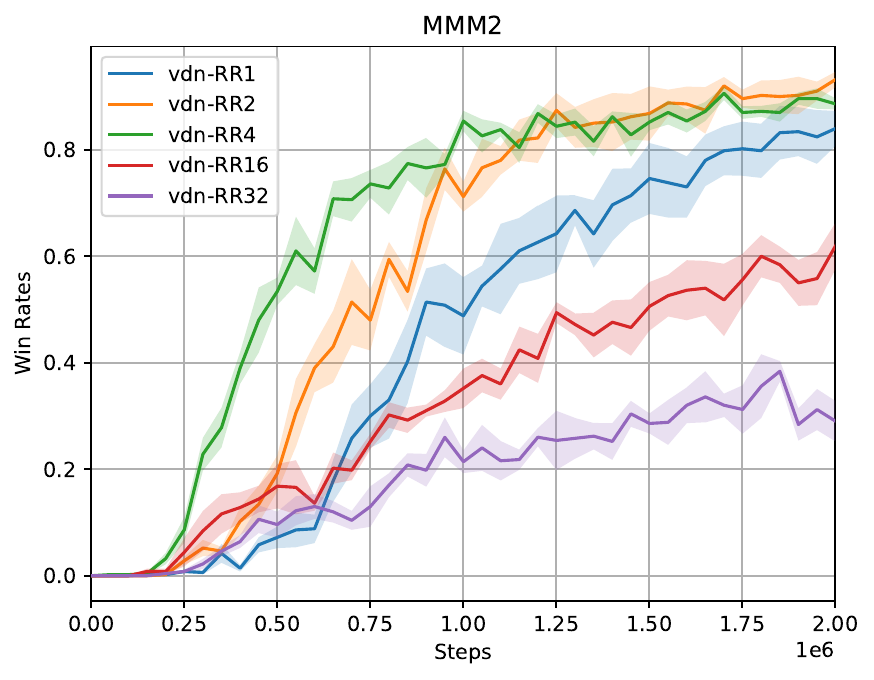}
    \caption{The performance of VDN on MMM2 task under different RR values. The results are plotted with standard errors among 5 random seeds.}
    \label{fig:tune_rr}
\end{figure}

Before we formally introduce our methods, we first define the agents. By using a shared neural network among different agents, we define the local utility function as $u_i = u_{\theta}(o_i, a_i)$. The policy can be derived from this utility function $\pi_i := \arg\max_{a_i} u_{\theta}(o_i, a_i)$. We define the joint utility as $\mathbf{u} = [u_i]_{i=0}^n$. To learn with the global reward, we define a mixer function $g_{\phi}(s, \mathbf{u})$. To obtain more stable training, we defined the target functions $u_{\hat{\theta}}(o_i, a_i)$ and $g_{\hat{\phi}}(s, \mathbf{u})$. The TD loss is calculated as
\begin{align}
    \mathcal{L}_{\theta} = \mathbb{E}_{\tau\sim \mathcal{B}} \mathbb{E}_{T\sim \tau}
    \big[g_{\phi}(s, \mathbf{u_{\theta}}) - \big(r^{tot} + \gamma g_{\hat{\phi}}(s', \mathbf{u_{\hat{\theta}}})\big) \big]^2,
\end{align}
and the agent parameter is updated by applying the gradient descent operator
\begin{align}
    \theta_{t+1} &= \theta_{t} - \alpha_\theta \nabla \mathcal{L}_{\theta},~\label{eqn:utilit_gradient_step}\\
    \phi_{t+1} &= \phi_{t} - \alpha_\phi \nabla \mathcal{L}_{\phi},~\label{eqn:mixer_gradient_step}
\end{align}
where $\alpha_\theta$ and $\alpha_\phi$ are the learning rates of $\mathbf{u_{\theta}}$ and $g_{\phi}$, respectively. The target functions are updated by exponential moving average with a proportion $\eta \in (0, 1.0)$.

As shown in Figure~\ref{fig:demo}, we use $N$-gradient backpropagation as a high RR.
The multiple gradient steps are made and the agent parameter $\theta_t$ becomes $\theta_{t+N}$ after one training step to better exploit the collected data in the buffer $\mathcal{B}$.  
We summarize the pseudo-code in Algorithm~\ref{alg:alg1}.

In the following subsections, we introduce several design options for using higher RRs in MARL.
\begin{algorithm}[t]
\caption{MARL Training}
\label{alg:alg1}
\begin{algorithmic}[1]
    \STATE \textbf{Hyperparameters:} replay ratio $N$, learning rate $\alpha_{\theta}$ and $\alpha_{\phi}$, $\eta_{\theta}$, $\eta_{\alpha}$
    \STATE \textbf{Initialize:} Environment $\mathcal{E}$, $\theta$, $\theta'$, $\phi$ , $\phi'$, $\mathcal{D}=\emptyset$
    \FOR{$j=1, 2, \cdots, J$}
        \STATE $\tau_j \sim \mathcal{E}$
        \STATE $\mathcal{D} =  \mathcal{D} \cup \{\tau_j\}$
        \FOR{$k=1, 2, \cdots, N$}
            \STATE $\mathcal{B}_k \sim \mathcal{D}$
            \STATE $\theta \leftarrow \theta - \alpha_{\theta} \nabla\mathcal{L}_{\theta}\label{alg:q_tilde}(\theta, \mathcal{B}_k)$ (Equation~\ref{eqn:utilit_gradient_step})
            \STATE $\phi \leftarrow \phi - \alpha_{\phi} \nabla\mathcal{L}_{\phi}\label{alg:q_tilde2}(\phi, \mathcal{B}_k)$ (Equation~\ref{eqn:mixer_gradient_step})
            \STATE $\theta' \leftarrow (1-\eta_{\theta})\theta' + \eta_{\theta} \theta$
            \STATE $\phi' \leftarrow (1-\eta_{\alpha})\phi' + \eta_{\alpha} \phi$
        \ENDFOR
    \ENDFOR
\end{algorithmic}
\end{algorithm}

\subsection{Batch Resampling}
At each training step, a data batch is sampled from the replay buffer for loss calculation. When $RR>1$, there is a choice of resampling a new batch for each loss calculation. Surprisingly, we find that multiple gradient updates with the same batch could still improve performance. However, there is a risk of overfitting to a minimum by using the same batch. Therefore, we resample the batch after each gradient update.

\subsection{Quantifying the risk of losing network plasticity}\label{sec:plasticity}
Using a $RR>1$ leads to a linear growth of the number of gradient steps. Recent studies in continual learning and single agent RL present that the plasticity of neural networks is lost with more gradient steps. The loss of plasticity raises issues for RL such as primacy bias~\citep{nikishin2022primacy}, inefficient exploration~\citep{xu2024drm}, etc. To measure the risk of plasticity loss in MARL, we use the Dormant Neural Ratio (DNR)~\citep{Sokar23Dormant} as an indicator of plasticity.
\begin{definition}
For the $\ell$-th layer of a neural network, let $h(x)^{\ell}$ denote the activation of all neurons and $h(x)^{\ell}_i$ denotes the $i$-th neuron. $x\in D$, where $D$ is the output of the last layer for all data in the batch. The score of the $i$-th neuron is defined as follows
\begin{align}
    d^{\ell}_i = \frac{\mathbb{E}_{x\in D}|h (x)^{\ell}_i|}{\frac{1}{H^{\ell}} \sum_{k\in h} \mathbb{E}_{x\in D} |h (x)^{\ell}_k|},
\end{align}
where $H^{\ell}$ is the output dimension of layer $\ell$. The $i$-th neuron in layer $\ell$ is $\rho$-dormant if $d^{\ell}_i \leq \rho$. The DNR for layer $\ell$ is defined as the proportion of the $\rho$-dormant neurons.
\end{definition}

Surprisingly, the DNR of the shared RNN neural network used in our MARL training maintains a low level of DNR during training, which indicates a low risk of plasticity loss. This observation leads to our simple method without extra mechanisms to address the plasticity loss.

By comparing the RNN agent and the non-RNN agent in MARL training (see Figure~\ref{fig:dormant_ratio}), we find the RNN might help maintain the network plasticity. For a feedforward neural network without RNN, it learns a static $X-Y$. Instead, the RNN network takes a latent variable from the previous time step and breaks the static $X-Y$ relation. That is, the output $\hat{Y}$ depends not only on $X$ but also on the latent variable $H$. The $H$, produced by the history trajectory, could result in different $\hat{Y}$ for the same $X$. This natural property of RNN increases the stochasticity during the network training, which might hinder the neurons from dormant.
\begin{figure}[t]
    \centering
    \begin{subfigure}[b]{0.24\textwidth}
        \centering
        \includegraphics[width=\textwidth]{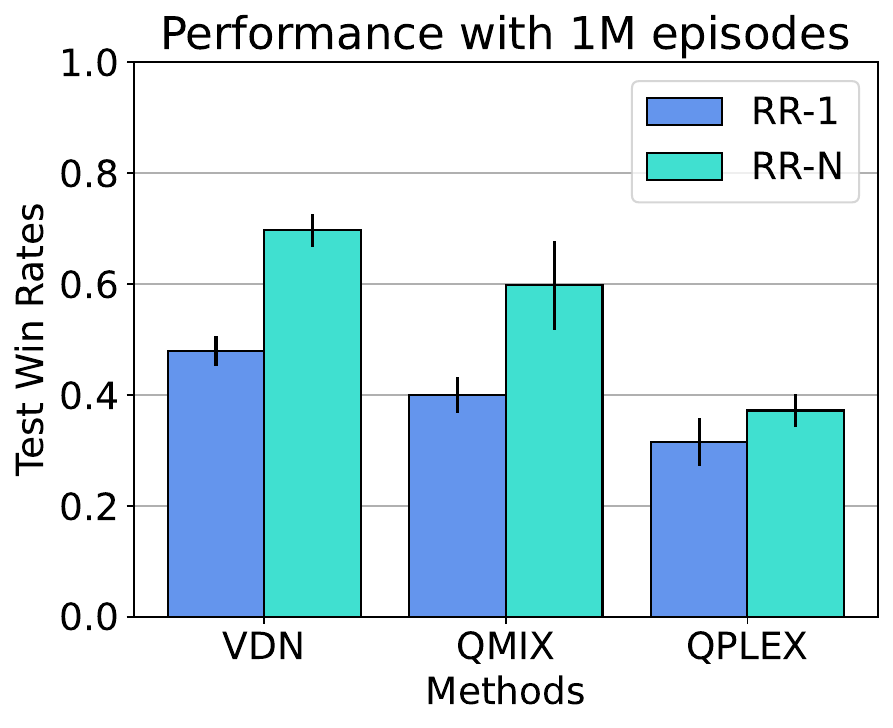}
        \label{fig:1m_performance}
     \end{subfigure}
    \begin{subfigure}[b]{0.24\textwidth}
        \centering
        \includegraphics[width=\textwidth]{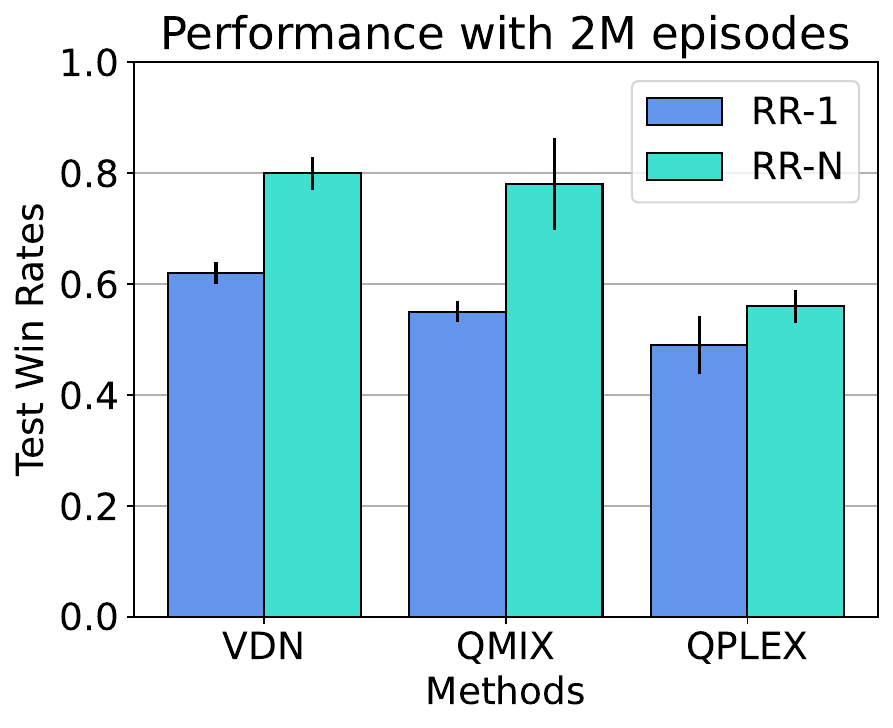}
        \label{fig:2m_performance}
    \end{subfigure}
        \vspace{-3mm}
    \caption{Comparison of common MARL training (i.e., $RR=1$) and using a higher RR (the best performance with $RR\in\{2, 4\}$) in $6$ Starcraft-II tasks. $3$ MARL methods are evaluated and their performances of using $1$ million and $2$ million environmental interactions are visualized.}\label{fig:1m_2m_performance}
\end{figure}

\section{Experiments}

In this section, we provide empirical analysis for $3$ widely investigated methods with higher RRs using mean $\pm$ standard error with five random seeds.
Our experiments focus on the following research questions~(RQs). \textbf{RQ1:} Does a higher RR increase the sample efficiency for MARL methods? \textbf{RQ2:} Does a higher RR raise the concern of losing network plasticity for MARL training?  \textbf{RQ3:} How does the computation budget (the number of updates) trade off with the environment interaction budget?

\subsection{Experimental setting}

\textbf{Baselines.} To show that a higher RR is a general approach to improve sample efficiency, we evaluate 3 value-based MARL methods: VDN~\citep{SunehagVDN}, QMIX~\citep{rashid2020monotonic}, and QPLEX~\citep{wang2021qplex}. The 3 methods aim to learn a local utility function $u_i(o_i, a_i)$ for the agent $i$. The Q-value function $g(s, \mathbf{u})$ is a composition of the local utility function. VDN defines an additive Q-value function $g(\mathbf{u}) = \sum_i u_i$. QMIX defines a monotonic Q-value function that satisfies $\frac{\partial g}{\partial u_i} \geq 0, \forall i \in \{1,...,n\}$. The last one, QPLEX utilizes a dueling structure based on the additive Q-value function to enhance the expressiveness.

\textbf{Environment.} For our performance evaluation, we select 6 tasks with different difficulties from the SMAC benchmark~\citep{samvelyan2019starcraft}. The SMAC benchmark is built on StarCraft II for cooperative game study. We select two easy tasks: \textit{2s\_vs\_1sc}, where two Stalkers are controlled to play against $1$ Spine Crawler; \textit{3s\_5z} where each side controls $3$ Stalkers and $5$ Zealots. We also select four super hard tasks: \textit{MMM2} where each group includes $1$ Medivac, $2$ Marauders and $7$ Marines; \textit{3s\_vs\_5z} where the agent controls $3$ Stalkers to play against $5$ Zealots, \textit{3s5z\_vs\_3s6z} with $3$ Stalkers and $5$ Zealots to play against $3$ Stalkers and $6$ Zealots, and \textit{corridor} where $6$ Zealots play against $24$ Zerglings.

\begin{figure*}[h]
     \centering
     \begin{minipage}{0.9\linewidth}
        \centering
         \begin{subfigure}[t]{0.3\textwidth}
             \centering
             \includegraphics[width=\textwidth]{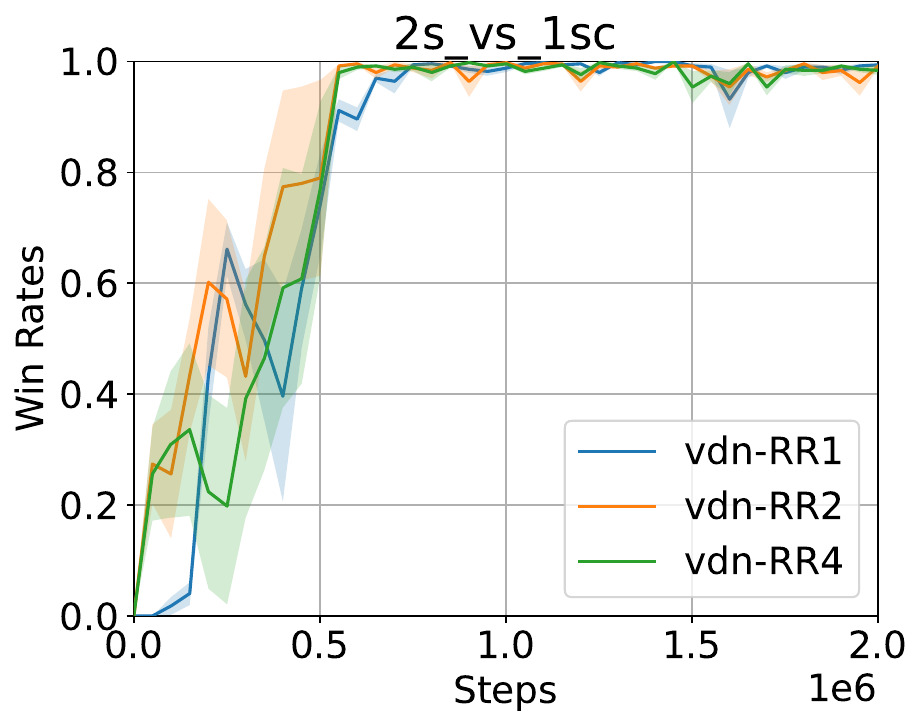}
             \label{fig:s11}
         \end{subfigure}
         \begin{subfigure}[t]{0.3\textwidth}
             \centering
             \includegraphics[width=\textwidth]{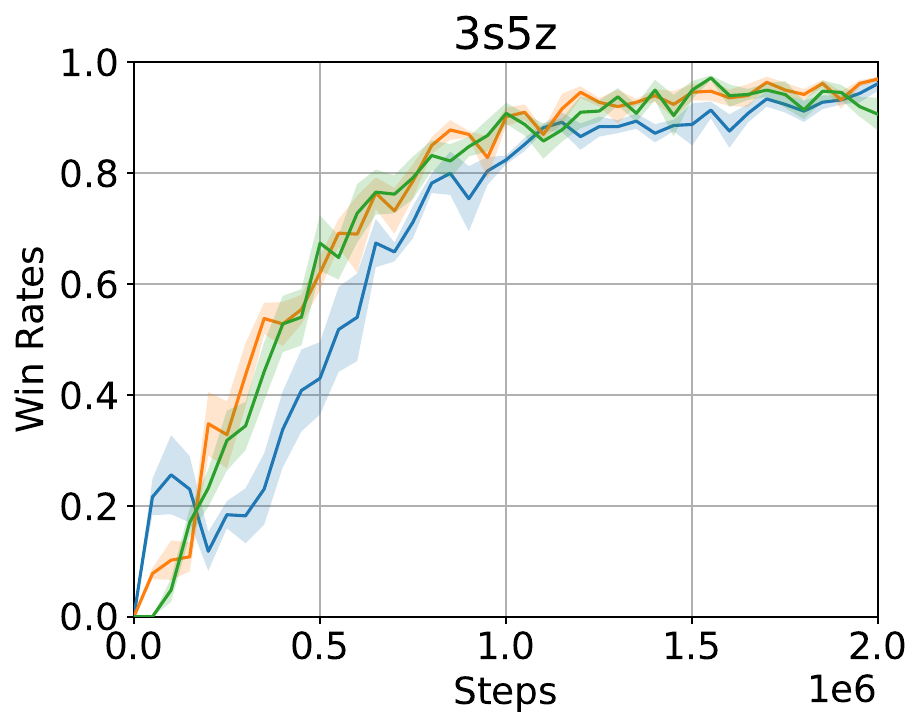}
             \label{fig:s1asd2}
         \end{subfigure}
         \begin{subfigure}[t]{0.3\textwidth}
             \centering
             \includegraphics[width=\textwidth]{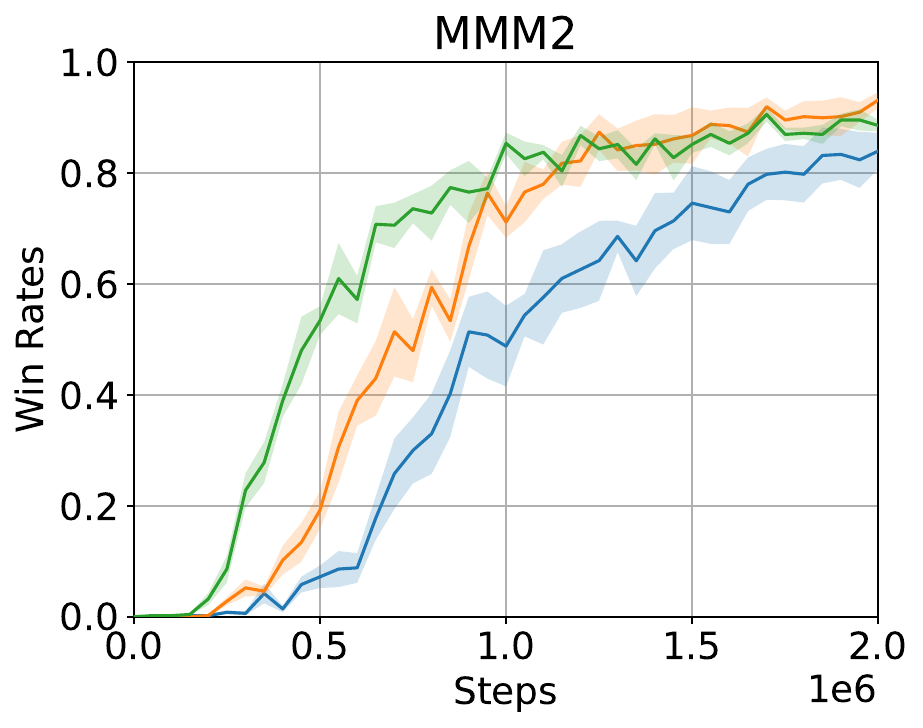}
             \label{fig:sfa133}
         \end{subfigure}
         \vspace{-3mm}\\ 
        \begin{subfigure}[b]{0.3\textwidth}
             \centering
             \includegraphics[width=\textwidth]{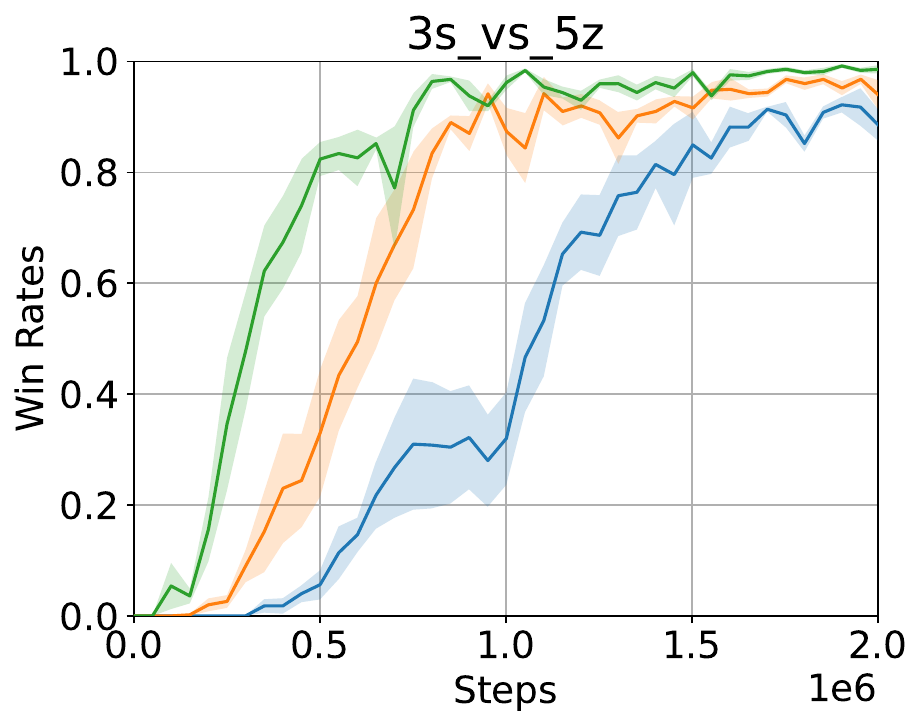}
         \end{subfigure}
          \begin{subfigure}[b]{0.3\textwidth}
             \centering
             \includegraphics[width=\textwidth]{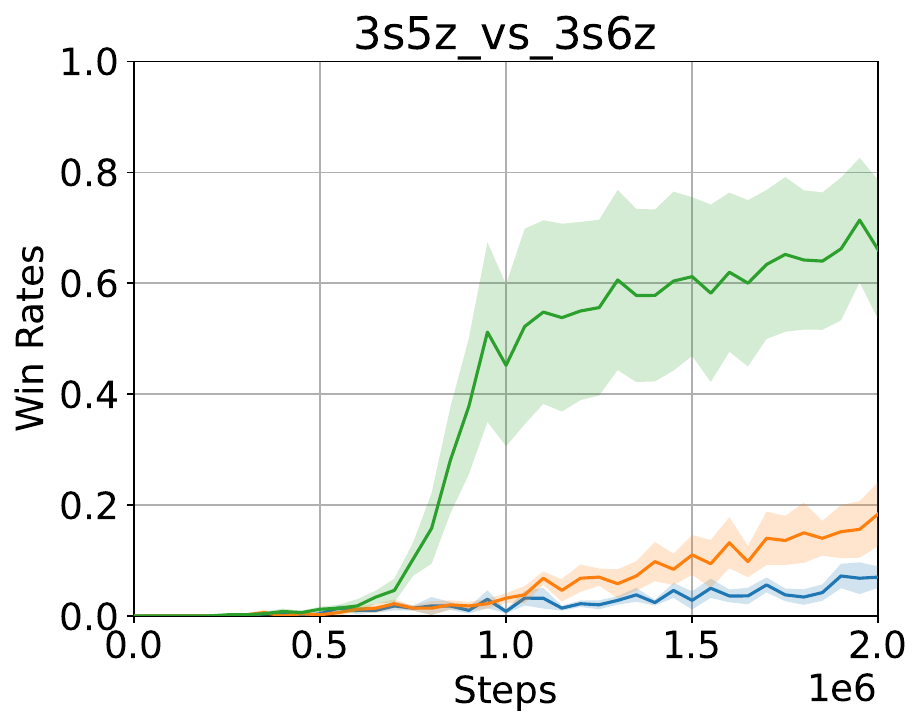}
         \end{subfigure}
         \begin{subfigure}[b]{0.3\textwidth}
             \centering
             \includegraphics[width=\textwidth]{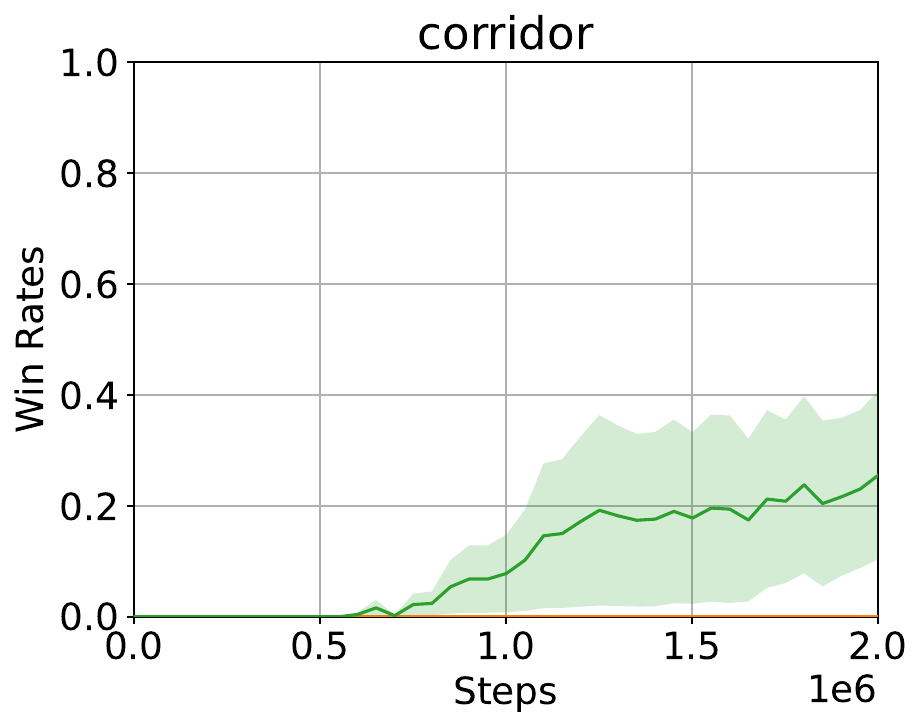}
         \end{subfigure}
     \end{minipage}
     \vspace{-1mm}
    \caption{The evaluation performances with different checkpoints from the VDN training.}\label{fig:vdn_performance}
    \vspace{1mm}
    \begin{minipage}{0.9\linewidth}
        \centering
         \begin{subfigure}[t]{0.3\textwidth}
             \centering
             \includegraphics[width=\textwidth]{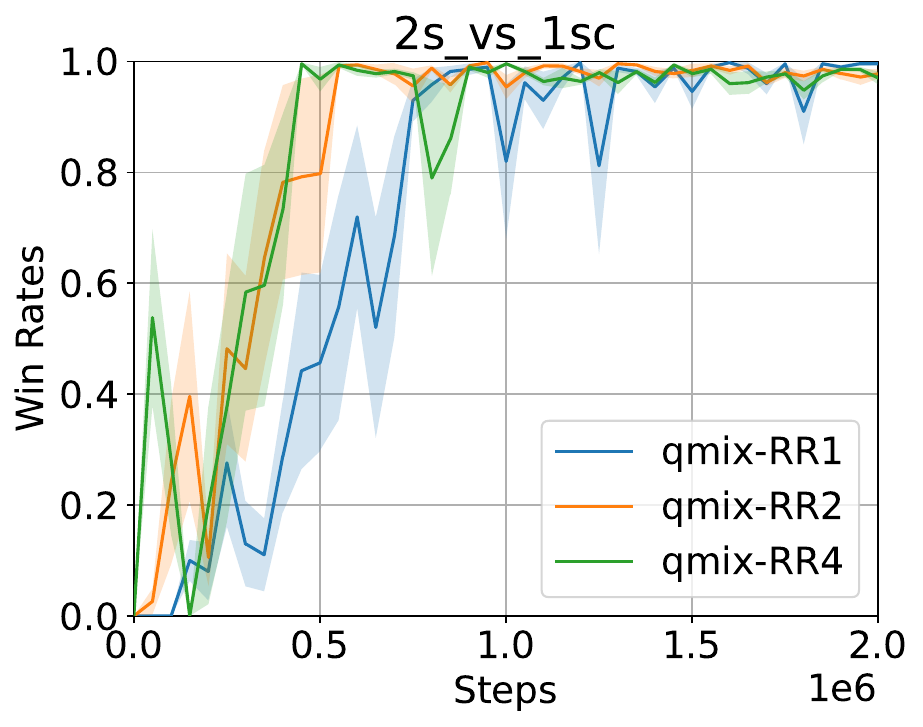}
             \label{fig:s12ga3}
         \end{subfigure}
         \begin{subfigure}[t]{0.3\textwidth}
             \centering
             \includegraphics[width=\textwidth]{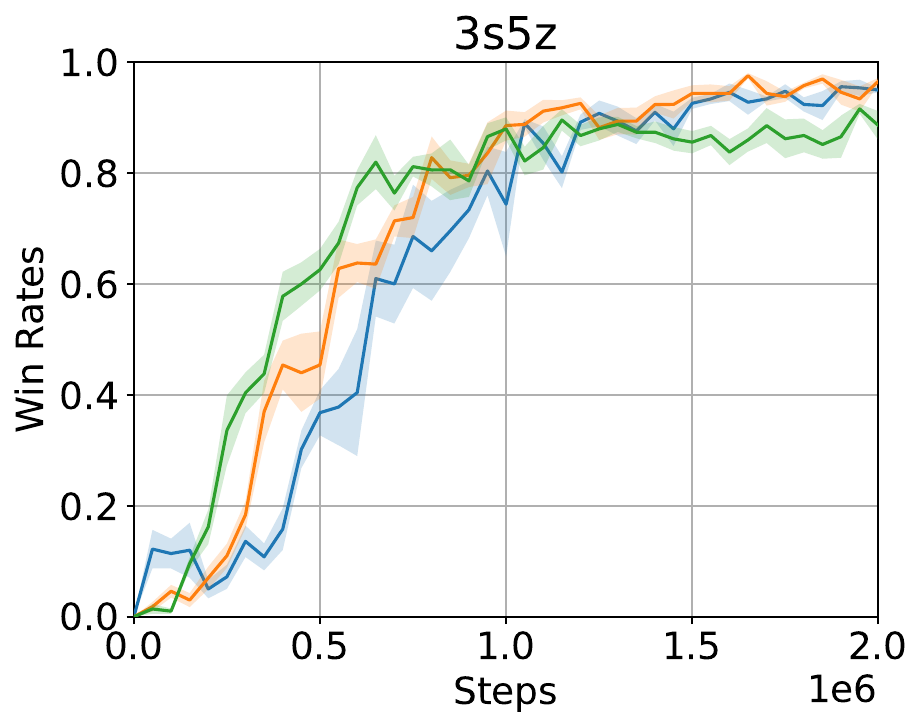}
             \label{fig:s1234}
         \end{subfigure}
         \begin{subfigure}[t]{0.3\textwidth}
             \centering
             \includegraphics[width=\textwidth]{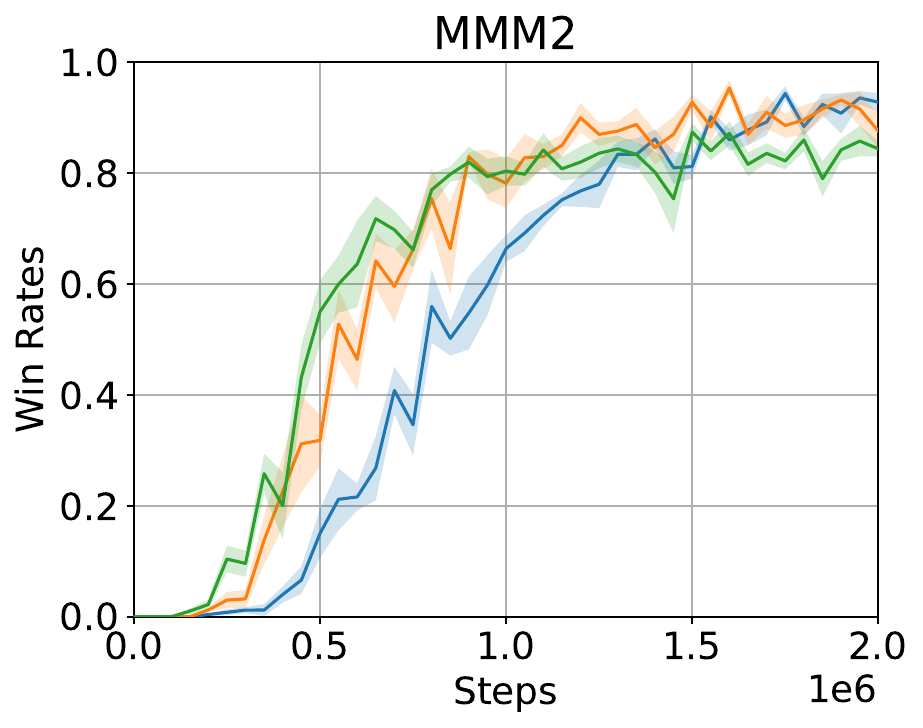}
             \label{fig:szgx13}
         \end{subfigure}
         \vspace{-3mm}\\
        \begin{subfigure}[b]{0.3\textwidth}
             \centering
             \includegraphics[width=\textwidth]{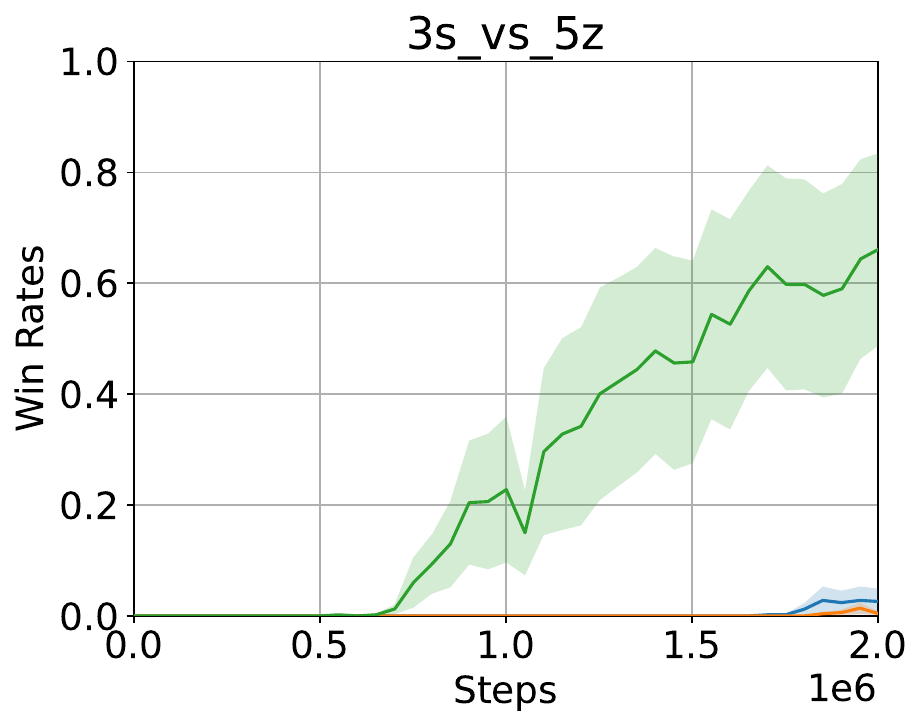}
         \end{subfigure}
          \begin{subfigure}[b]{0.3\textwidth}
             \centering
             \includegraphics[width=\textwidth]{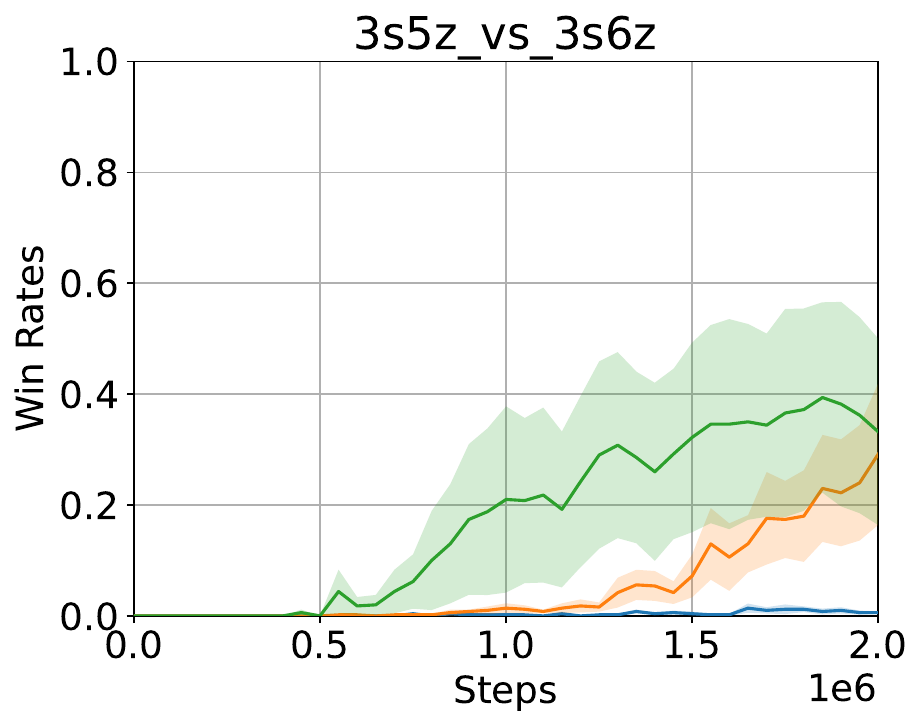}
         \end{subfigure}
         \begin{subfigure}[b]{0.3\textwidth}
             \centering
             \includegraphics[width=\textwidth]{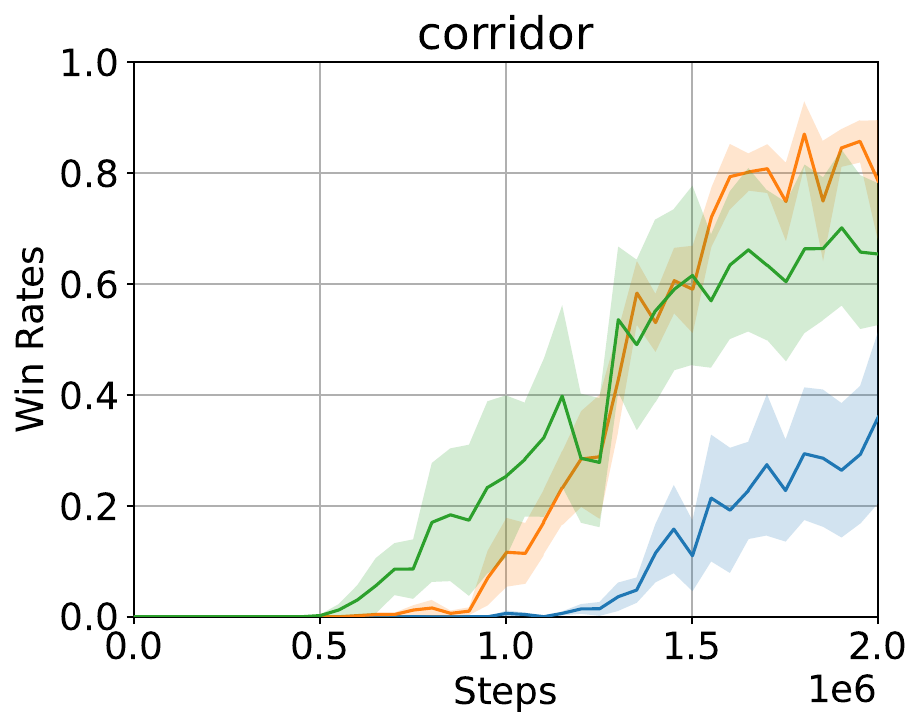}
         \end{subfigure}
     \end{minipage}
     \vspace{-1mm}
    \caption{The evaluation performances with different checkpoints from the QMIX training.}\label{fig:qmix_performance}
    \vspace{-1mm}
\end{figure*}
\begin{figure*}[h]
     \centering
     \begin{minipage}{0.9\linewidth}
        \centering
         \begin{subfigure}[t]{0.3\textwidth}
             \centering
             \includegraphics[width=\textwidth]{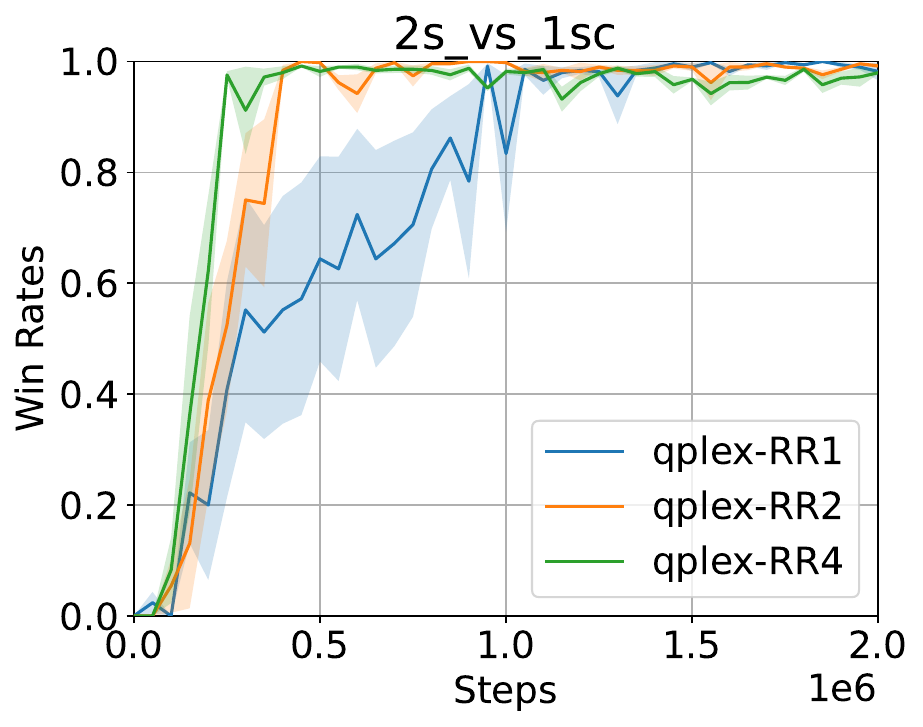}
             \label{fig:ssaf13}
         \end{subfigure}
         \begin{subfigure}[t]{0.3\textwidth}
             \centering
             \includegraphics[width=\textwidth]{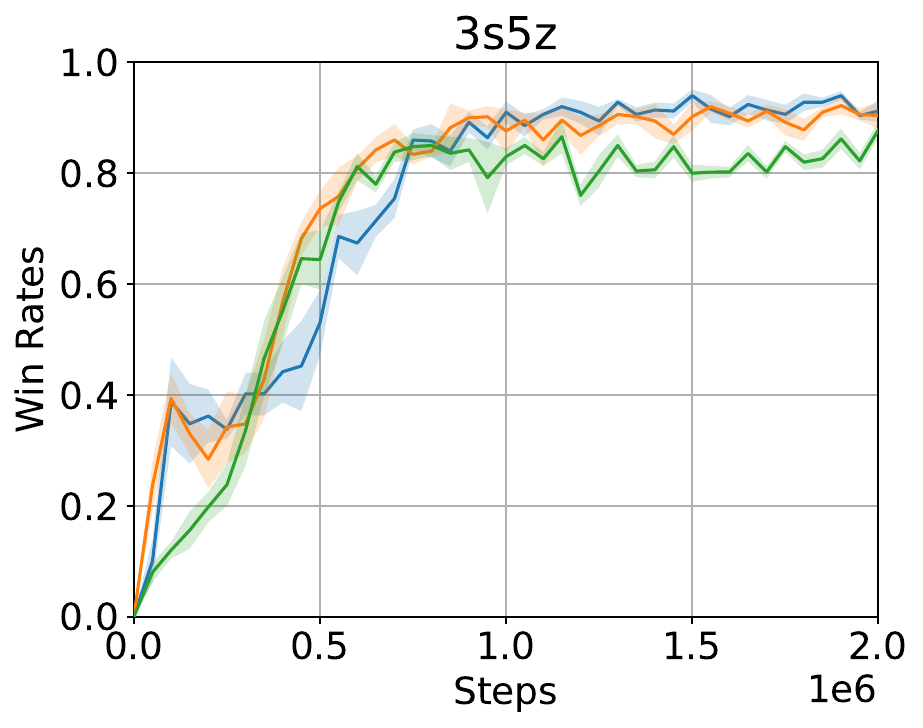}
             \label{fig:s12}
         \end{subfigure}
         \begin{subfigure}[t]{0.3\textwidth}
             \centering
             \includegraphics[width=\textwidth]{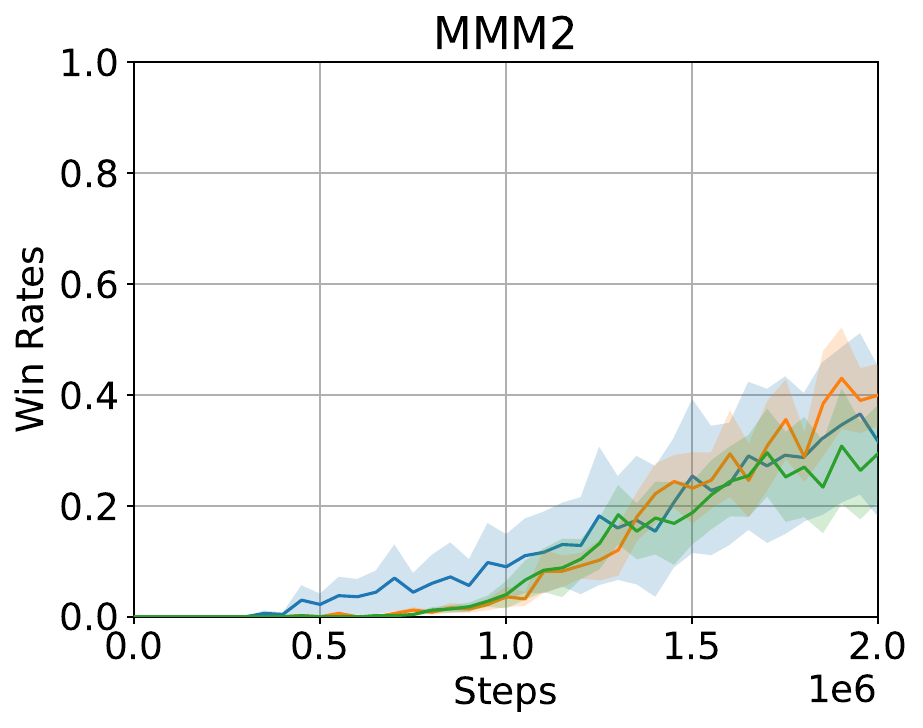}
             \label{fig:s133131}
         \end{subfigure}
        \begin{subfigure}[b]{0.3\textwidth}
             \centering
             \includegraphics[width=\textwidth]{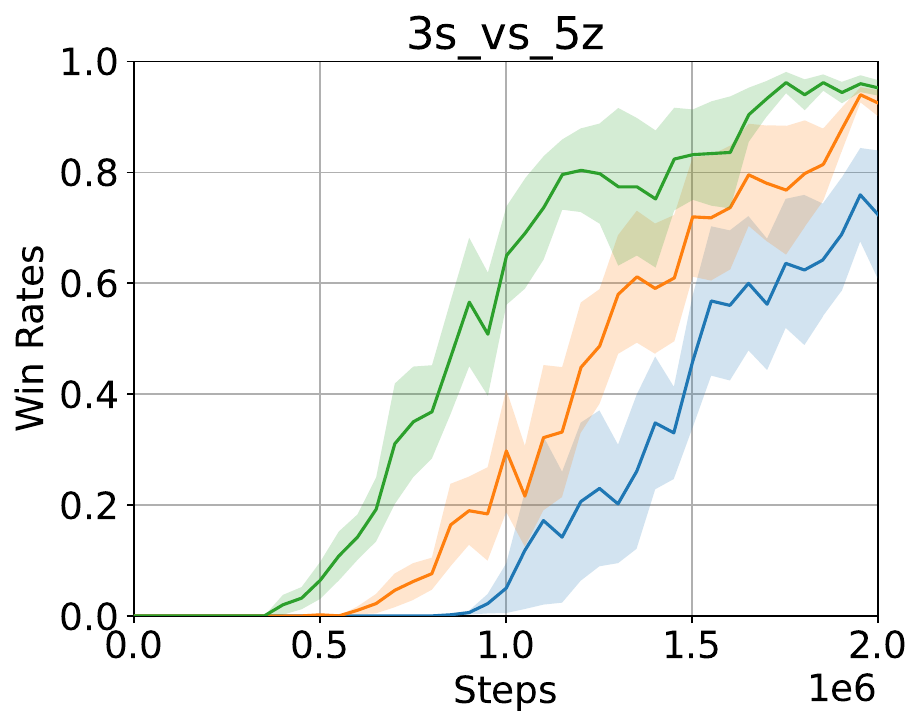}
         \end{subfigure}
          \begin{subfigure}[b]{0.3\textwidth}
             \centering
             \includegraphics[width=\textwidth]{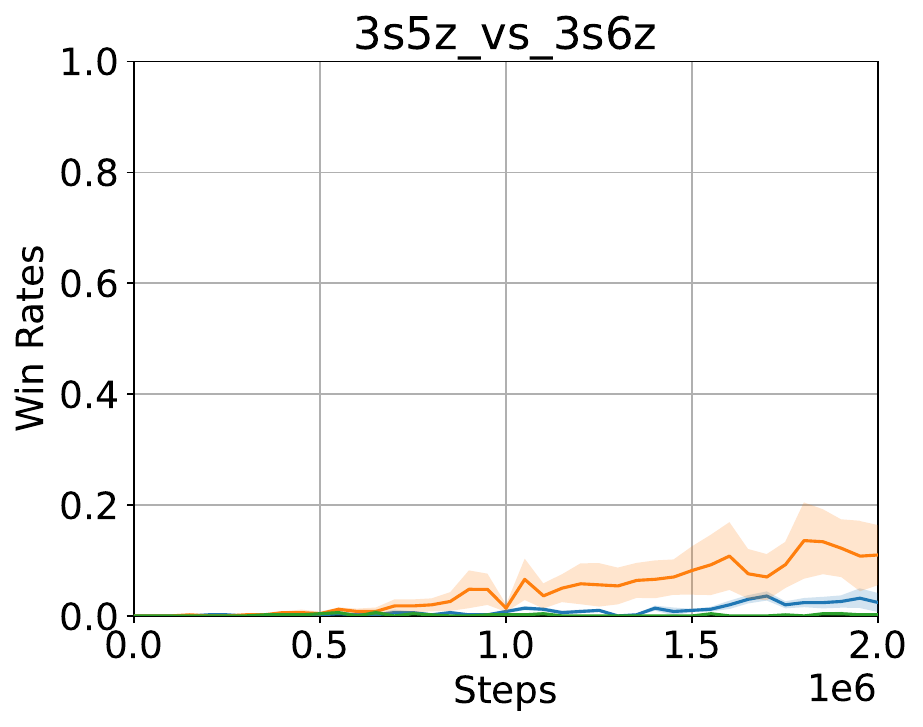}
         \end{subfigure}
         \begin{subfigure}[b]{0.3\textwidth}
             \centering
             \includegraphics[width=\textwidth]{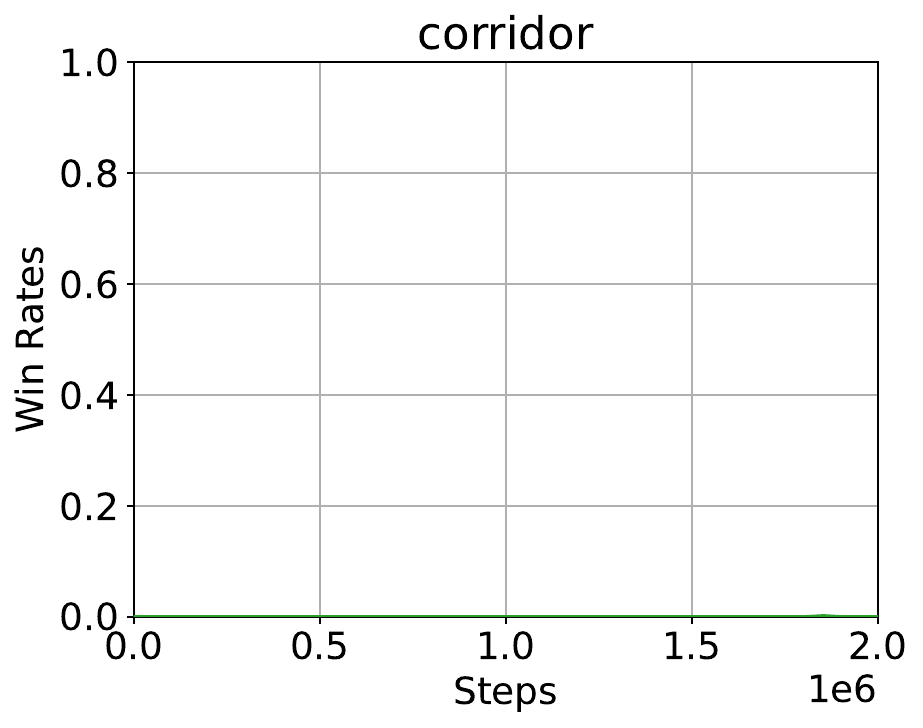}
         \end{subfigure}
     \end{minipage}
    \caption{The evaluation performances with different checkpoints from the QPLEX training.}
        \label{fig:qplex_performance}
\end{figure*}

\subsection{\textbf{RQ1:} The effect of high RRs}~\label{sec:performance}
To reveal the effect of $RR > 1$, we first visualize the performance of VDN under different $RR$ values on the \textit{MMM2} task. As presented in Figure~\ref{fig:tune_rr}, we observed that $RR\in\{ 2, 4 \}$ could drastically improve the sample efficiency but larger RRs cause poor performance. A possible reason for the performance decrease is that exploiting the collected data too frequently causes the overfitting of current data and thus the agent soon fails to efficiently explore the environment. In conclusion, the results verify that the current MARL training does not efficiently exploit the collected data.

Based on the above observation, we apply higher RR(s) on 3 baseline methods: VDN, QMIX, and QPLEX. In Figure~\ref{fig:1m_2m_performance}, we present the averaged performance on 6 tasks for each agent. To show the effectiveness of $RR>1$, we tune $RR\in\{2, 4\}$ and select the best performance task-wise to calculate the averaged performance, the $RR-N$ agent resulting in Figure~\ref{fig:1m_2m_performance}. From the results, we observe that using RRs higher than the base-rate $1$ significantly improves the agents' win rates. Furthermore, this improvement is consistent for all the $3$ MARL methods with both $1$ million episodes and $2$ million training episodes.

For each MARL method, the detailed results can be found in Figure~\ref{fig:vdn_performance}, \ref{fig:qmix_performance} and~\ref{fig:qplex_performance}. For VDN, using $RR=2, RR=3$ improves the sample efficiency in $5$ tasks and $6$ tasks, respectively. In super hard tasks, using $RR=4$ achieves much higher win rates as well as faster convergence compared to the base-rate $1$. For QMIX, $RR=4$ improves the sample efficiency in all $6$ tasks and $RR=2$ does the same except for the \textit{3s\_vs\_5z} task. Using $RR=4$, although the performance converges with fewer samples, the performance seems to converge to a local minimum in tasks \textit{3s5z} and \textit{MMM2}, indicating an overfit on the previously collected data. Nevertheless, $RR=4$ achieves faster convergence and higher win rates in other tasks. For QPLEX, the $RR=2$ achieves faster convergences in $3$ of the $6$ tasks: \textit{2s\_vs\_1sc}, \textit{3s\_vs\_5z} and \textit{3s5z\_vs\_3s6z}. In \textit{MMM2}, $RR=2$ converges slower than the base-rate $1$ but achieves a higher final win rate. $RR=4$, even though converge faster in \textit{2s\_vs\_1sc, 3s\_vs\_5z}, fails to outperforms the base-rate $1$ in other tasks.

In conclusion, we find the improvement in sample efficiency is consistent for all 3 MARL methods. It is worth noting that on some super hard tasks such as \textit{3s\_vs\_5z, 3s\_vs\_3s6z} and \textit{corridor}, where the $RR=1$ fails to converge to a satisfying win rate, $RR>1$ can still achieve competitive win rates. In these tasks, the original methods with $RR=1$ achieve only near zero win rates (VDN in \textit{3s5z\_vs\_3s6z} and \textit{corridor}, QMIX in \textit{3s\_vs\_5z} and \textit{3s5z\_vs\_3s6z}). Surprisingly, by only increasing the RR, the agent achieves satisfying win rates within $2$ million time steps, which indicates the effectiveness of using a higher RR.

\subsection{\textbf{RQ2:} Higher RR and plasticity loosing}
\begin{figure*}[h]
     \centering
     \begin{subfigure}[b]{0.32\textwidth}
         \centering
         \includegraphics[width=0.9\textwidth]{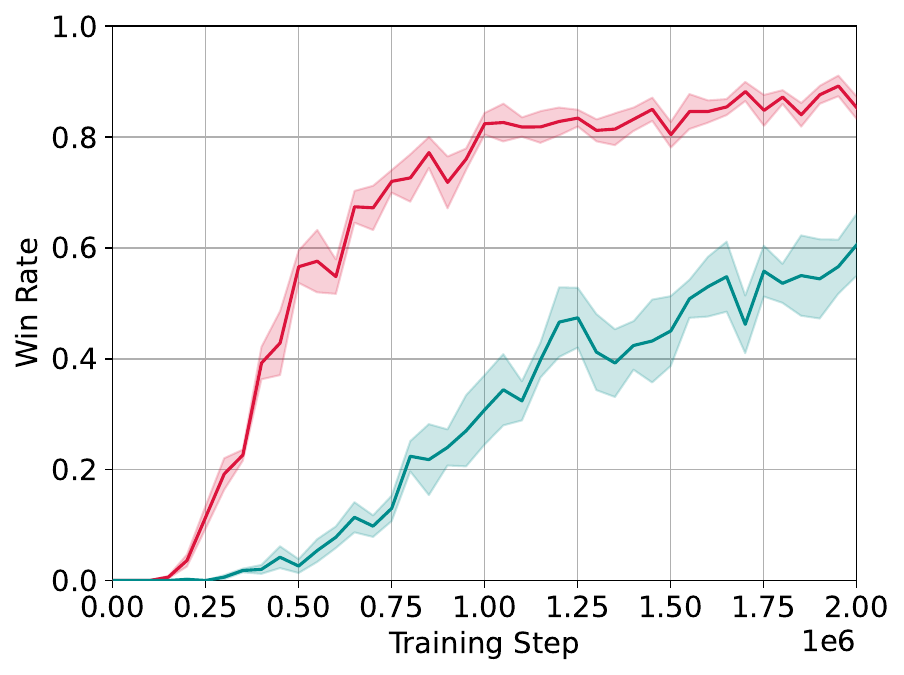}
         \caption{}
         \label{fig:performance_non_rnn}
     \end{subfigure}
     \begin{subfigure}[b]{0.32\textwidth}
         \centering
         \includegraphics[width=0.9\textwidth]{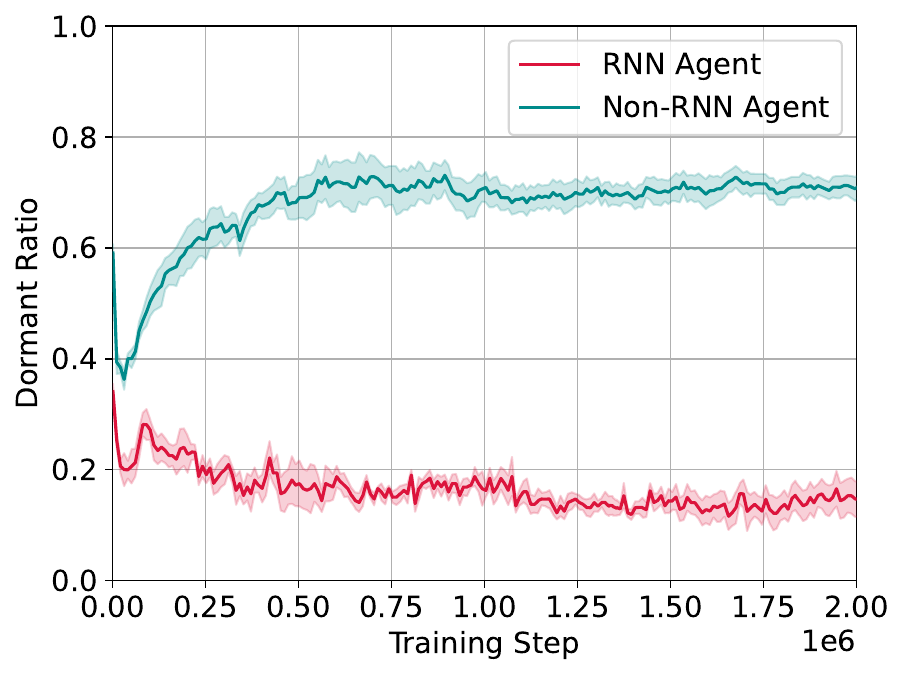}
         \caption{}
         \label{fig:dormant_ratio}
     \end{subfigure}
     \begin{subfigure}[b]{0.32\textwidth}
         \centering
         \includegraphics[width=0.9\textwidth]{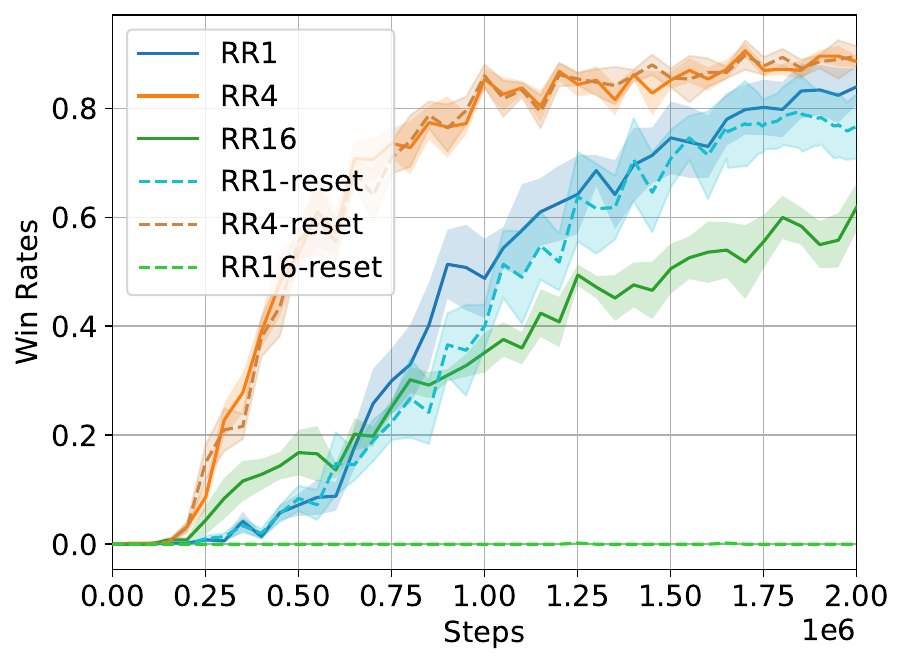}
         \caption{}
         \label{fig:performance_reset}
     \end{subfigure}
        \caption{\textbf{Left}: Performances on the MMM2 task. \textbf{Middle}: Dormant ratios during the training. \textbf{Right}: Performances of the agents with or w/o reset.}
        \label{fig:plasticity}
\end{figure*}

Using higher RR values increases the number of parameter updates linearly. Hence using larger RRs potentially increases the risk of losing the network plasticity. In this subsection, we empirically analyze the plasticity loss in MARL training.

In Figure~\ref{fig:plasticity}, we visualize the DNR with $\rho=0.001$ (See Section~\ref{sec:plasticity}) for two agents, one with an RNN layer and another one consisting of only fully-connected layers. For non-RNN agents, we observe an increasing DNR in the beginning and then the DNR is maintained at a high level. Surprisingly, the RNN agent succeeds in maintaining a low-level DNR during the training.

To further verify the plasticity of the RNN agent, we apply the reset operator, which is widely used with higher RR to address network plasticity issues. The performances of different training checkpoints in \textit{MMM2} are visualized in Figure~\ref{fig:performance_reset}. From the results, we see no significant performance change between using resetting or not using it under small RR values ($1$ and $4$). However, when using a large RR value ($16$ in our experiment), using reset causes performance collapse. These empirical results confirm that the RNN agent naturally maintains good network plasticity, hence the extra technique is not effective here.

\subsection{\textbf{RQ3:} The trade-off between computation budgets and environment interaction budgets}

Although the previous experiments have shown that a higher RR improves the training sample efficiency for multiple MARL methods, the required computation is also increased. In this subsection, we visualize the agent performances among different computation budgets and data budgets. The computation budget is quantified with the number of parameter updates and the data budget is presented by the environment steps. The selected computation budgets are $1M, 2M, 4M,$ and $8M$ and the data budget is set to $0.5M, 1M$ and $2M$.

\begin{figure}
    \centering
    \includegraphics[width=0.42\textwidth]{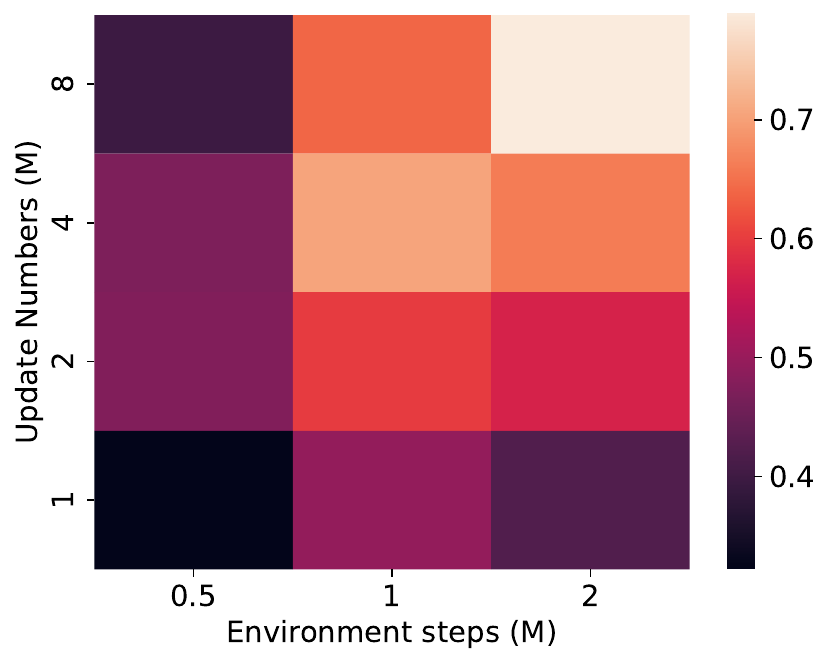}
    \caption{The averaged performance (win rates) of QMIX on $6$ SMAC tasks with different computation budgets and environmental interaction budgets.}
    \label{fig:budget}
\end{figure}

Figure~\ref{fig:budget} visualizes the average win rates for the QMIX agent on $6$ SMAC tasks that are used in Section~\ref{sec:performance}. In general, we find that higher RRs or more environment interactions both increase the win rates. By comparing the win rate increasing (with more update numbers) under different data budgets, we find that a larger data budget enables a more drastic win rate increase. With a low data budget ($0.5M$ in this case), the win rate only increases slightly.

\subsection{Empirical analysis of other hyper-parameters}~\label{sec:lr_bs}
Using a higher RR, each training step exploits more samples and updates the network parameters faster. One might ask whether only using more samples or only faster updates leads to performance improvement. To answer this question, we investigate the effect of using larger batch sizes or using higher learning rates.

\textbf{Comparison to larger batch size:} Compared to the conventional training configuration for MARL methods, using higher RR means more samples (batch size $\times$ RR) are used in each training step. One might ask whether it is similar to a larger batch size. We train VDN agents in \textit{3s5z\_vs\_3s6z} task with batch size in $\{ 32, 64, 128 \}$ (default $32$) and compares to $RR=\{ 2, 4 \}$ with batch size $32$.

\begin{figure}[t]
    \centering
    \begin{subfigure}[b]{0.24\textwidth}
        \centering
        \includegraphics[width=\textwidth]{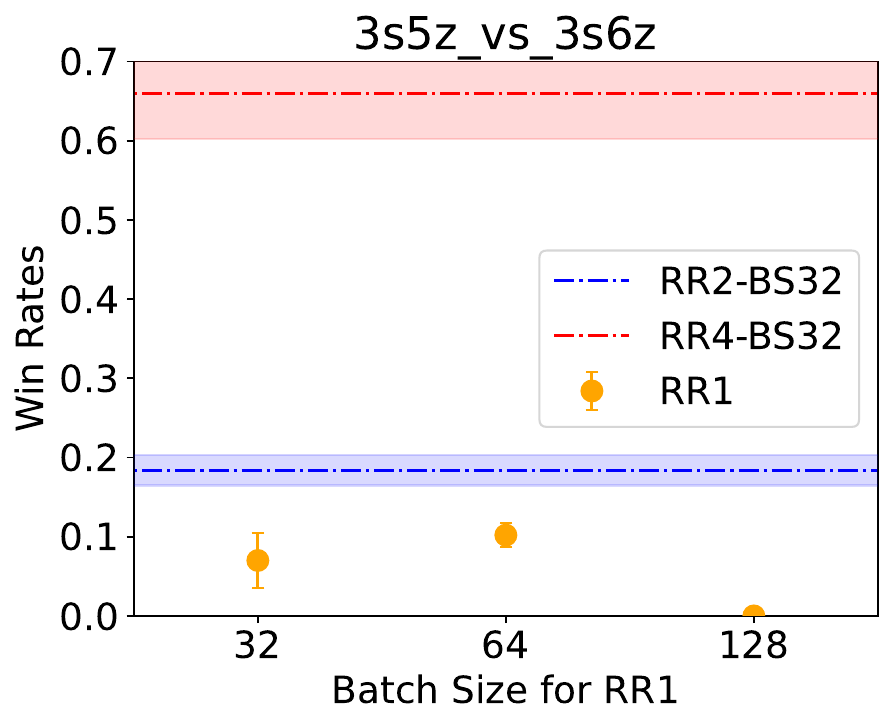}
        \label{fig:bs}
     \end{subfigure}
    \begin{subfigure}[b]{0.24\textwidth}
        \centering
        \includegraphics[width=\textwidth]{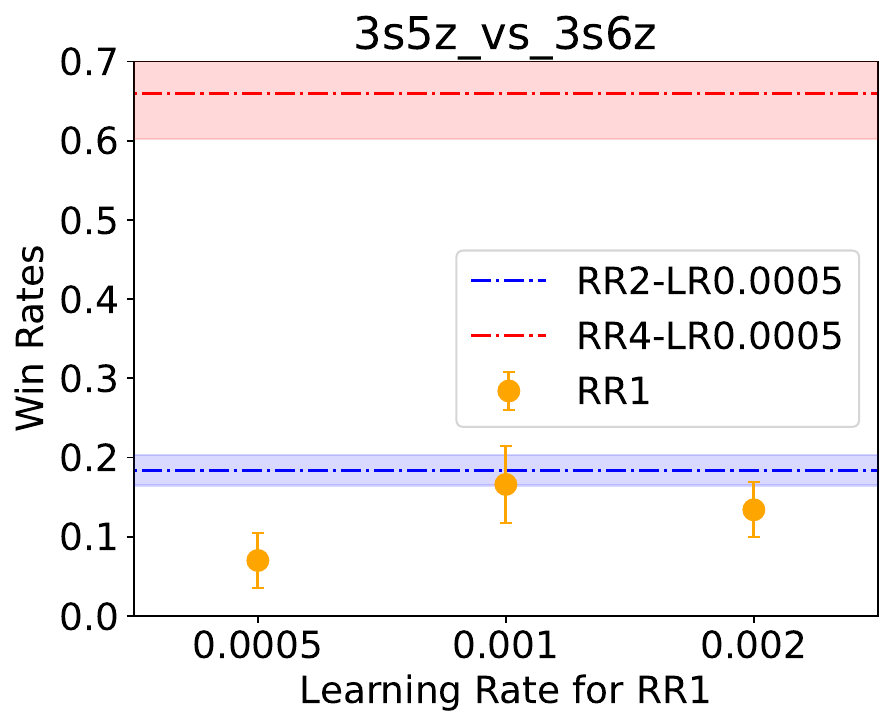}
        \label{fig:lr}
    \end{subfigure}
    \caption{Comparison with different batch sizes and learning rates. Both the shaded region and the error bars refer to standard errors.} \label{fig:bs_lr_performance}
    \vspace{-2mm}
\end{figure}

As seen from the results (Figure~\ref{fig:bs_lr_performance}), a batch size $64$ slightly improves the final win rate but a batch size $128$ causes the performance to collapse. Compared to the performance improvement brought by a slightly larger batch size, the $RR\in\{2, 4\}$ improves the performance more significantly.

\textbf{Comparison to a higher learning rate:} As higher RR takes several parameter updates for each training step, the parameter changes faster than that of $RR=1$. Similarly, increasing the learning rate may also accelerate the parameter updates. We compare the performance of higher RR and larger learning rates. We train VND agents with learning rates $\{ 0.0005, 0.001, 0.002\}$.

The results are presented in Figure~\ref{fig:bs_lr_performance}. We observe that using a learning rate from $\{0.001, 0.002\}$ improves the performance, where the learning rate $0.001$ achieves the same performance as $RR=2$. However, compared to $RR=4$, these improvements brought by a larger learning rate are less significant.

\section{Related Work}
To enhance the sample efficiency for MARL, several approaches have been developed. In value-based methods, one line of works~\citep{SunehagVDN, rashid2020monotonic, wang2021qplex, liu2023na2q} focuses on value decomposition, factorizing the global value function to simpler local utility function. These works developed different factorizing mechanisms to obtain efficient learning. Except for the commonly-used fully-shared neural network, Li et al.~\citep{li2021celebrating} separate the shared module and independent module to enhance the network capacity. 
Note that our contribution is not to propose new methods, but to modify widely used training mechanisms.

Higher RR is widely used in recent work. Nikishin et al.~\citep{nikishin2022primacy} propose to use a higher RR to enhance the training performance and address the primacy bias issue with reset. D'Oro et al.~\citep{d'oro2023sampleefficient} further raises the RR, using the shrink and perturbation instead of the reset to maintain the network plasticity. Built on a higher RR agent, Schwarzer et al.\citep{schwarzer2023bigger} proposed a model-free agent that achieves human-level performance on Atari games with 100k frames. Our work applies higher RR in the MARL domain, where the sample efficiency is more challenging.

\section{Conclusion and Future Work}
In this work, we propose to increase the RR to enhance the sample efficiency of MARL training. By investigating $3$ MARL methods on $6$ SMAC tasks, we show that simply using higher RRs can significantly improve the sample efficiency. As the previous study raises the concern of plasticity loss brought by higher RRs, we quantify the plasticity loss during MARL training. Our empirical results show that the use of RNNs helps to maintain network plasticity.

\textbf{Limitation}
The empirical results presented in this work have shown that higher RRs improve sample efficiency for MARL training. However, multiple gradient calculation at each training step increases the computation budget. This raises a trade-off between the computation and environmental interaction. For problems where the environmental interaction is risky or expensive, using a higher RR is an effective to avoid more environmental interaction thus reducing the risks and expense.

\textbf{Future Work} Although we have shown that increasing RR is an effective approach to enhance the sample efficiency for MARL training, higher RRs may be less effective in late training. Using adaptive learning rates or adaptive RR might solve this issue, we leave it for future work. Another open problem is that we only use relatively low RR values ($2$ and $4$), compared to tens of frames that are collected at each training step. It is an interesting direction to seek solutions that enable higher RR to further improve the sample efficiency. Another insight that this work provides is the inefficient training mechanism of MARL. Although we have shown that increasing RR could be one effective approach for better sample efficiency, investigating other methods to optimize the training process could enable us to better understand the sample efficiency problem for MARL.

\bibliographystyle{unsrt} 
\bibliography{conference_101719}

\end{document}